\documentclass[11pt]{article}
	
\usepackage[colorlinks=true, bookmarks=true]{hyperref}\usepackage{etoolbox}

\UseRawInputEncoding

\AtBeginDocument{%
  \hypersetup{
    citecolor=blue,
    linkcolor=red,   
    urlcolor=blue}
    }
	\newcommand{\blind}{0}
	
    \makeatletter
    \renewcommand\section{\@startsection {section}{1}{\z@}%
                                       {-3.5ex \@plus -1ex \@minus -.2ex}%
                                       {2.3ex \@plus.2ex}%
                                       {\normalfont\fontfamily{phv}\fontsize{16}{19}\bfseries}}
    \renewcommand\subsection{\@startsection{subsection}{2}{\z@}%
                                         {-3.25ex\@plus -1ex \@minus -.2ex}%
                                         {1.5ex \@plus .2ex}%
                                         {\normalfont\fontfamily{phv}\fontsize{14}{17}\bfseries}}
    \renewcommand\subsubsection{\@startsection{subsubsection}{3}{\z@}%
                                        {-3.25ex\@plus -1ex \@minus -.2ex}%
                                         {1.5ex \@plus .2ex}%
                                         {\normalfont\normalsize\fontfamily{phv}\fontsize{14}{17}\selectfont}}
    \makeatother
	
        \usepackage{setspace}
        \usepackage{amssymb}
        \usepackage{algorithm}
        \usepackage{algpseudocode}
	\usepackage{graphicx}
	\usepackage{enumerate}
	\usepackage{xcolor}
	\usepackage{natbib} 
    \usepackage{natbib} 
        \usepackage{url} 
        \usepackage{multirow}
        \usepackage{multicol}
        \usepackage{subcaption}
        \usepackage{nomencl}
        \makenomenclature
        \usepackage{etoolbox}
	\usepackage{url} 
        \usepackage{amsmath}
        \usepackage{graphicx}
        \usepackage{dsfont}
        \usepackage{enumerate}
        \usepackage{xcolor}
        \usepackage{booktabs}
        \renewcommand\nomgroup[1]{%
          \item[\bfseries
          \ifstrequal{#1}{P}{Abbreviations}{%
          \ifstrequal{#1}{N}{Notations}{%
          \ifstrequal{#1}{O}{Other symbols}{}}}%
        ]}
        
\begin{document}
		
\def\spacingset#1{\renewcommand{\baselinestretch}%
    {#1}\small\normalsize} \spacingset{1}
		
\if0\blind
{
    \title{Interval Prediction of Electricity Demand Using a Cluster-Based Block Bootstrapping Method}
    \author{Rohit Dube $^a$, Natarajan Gautam $^b$, \\ 
            Amarnath Banerjee $^a$, Harsha Nagarajan $^c$\\
            $^a$ Industrial and Systems Engineering, Texas A\&M University, College Station, USA \\
            $^b$ Electrical Engineering and Computer Science, Syracuse University, NY, USA \\
            $^c$ Applied Mathematics and Plasma Physics, Los Alamos National Laboratory, \\Los Alamos, USA}
    \date{}
    \maketitle
} \fi

\if1\blind
{
    \title{Interval Prediction of Electricity Demand Using a Cluster-Based Block Bootstrapping Method}
    \author{Author information is purposely removed for double-blind review}
    \maketitle
} \fi




            


\begin{abstract}
Accurate electricity demand prediction is critical for applications such as micro-grid operation, yet low levels of aggregation introduce large uncertainty that challenges traditional point forecasts. We propose a \emph{Cluster-based Block Bootstrapping}~(CBB) algorithm that forms prediction intervals by sampling residual blocks drawn from variance-homogeneous clusters identified via a neural-network spectral clustering step. Evaluated on smart-meter data from \(50\) households in Washington state, CBB \textbf{(i)} \emph{narrows the Winkler Score by up to \(22.6\,\%\) (and by \(10.7\,\%\) at the 90\,\% confidence level)} relative to ensemble \textbf{quantile-regression} baselines, while \textbf{(ii)} \emph{cutting training time by \(91.5\,\%\)} because only one point model is fitted. By aligning residual sampling with demand pattern similarity, clustering produces sharper intervals without sacrificing coverage, giving micro-grid operators fast and reliable uncertainty estimates without repeatedly training large model ensembles which is an important advancement for real-time decision-making under volatile demand.
\end{abstract}
			
	\noindent%
\newpage

	\spacingset{1.5} 
\nomenclature[P]{\(\text{MAE}\)}{Mean Absolute Error}
\nomenclature[P]{\(\text{MSE}\)}{Mean Squared Error}
\nomenclature[P]{\(\text{RMSE}\)}{Root Mean Squared Error}
\nomenclature[P]{\(\text{MAPE}\)}{Mean Absolute Percentage Error}
\nomenclature[P]{\(\text{R2}\)}{R Squared Error}
\nomenclature[P]{\(\text{RMSLE}\)}{Root Mean Squared Logarithm Error}

\printnomenclature

\section{Introduction} \label{s:intro}
The past decades have seen the emergence of deregulated electricity markets, where Independent System Operators (ISOs) facilitate the buying and selling of electricity. These ISOs conduct short-term market settlement of electricity prices for supply and demand generally at two time scales: day-ahead~(\(24-32\) hours) and real-time~(\(3-1\) hour) \cite{sioshansi2013evolution,stoft2002power}. A small residential Micro-Grid (MG) capable of local generation has been envisioned to participate in these electricity markets to reduce the load on the main grid, especially during peak demand periods. However, effective participation requires highly accurate short-term demand forecasts, particularly challenging in the low-aggregation setting of a small MG \cite{HIRSCH2018402, SOSHINSKAYA2014659}. The electricity consumption patterns exhibit pronounced uncertainty in such MGs due to factors like distributed energy generation, newer loads like electric vehicles and smart appliances, and dependence on weather.
\begin{figure}[htbp]
  \centering
    \includegraphics[scale=0.85]{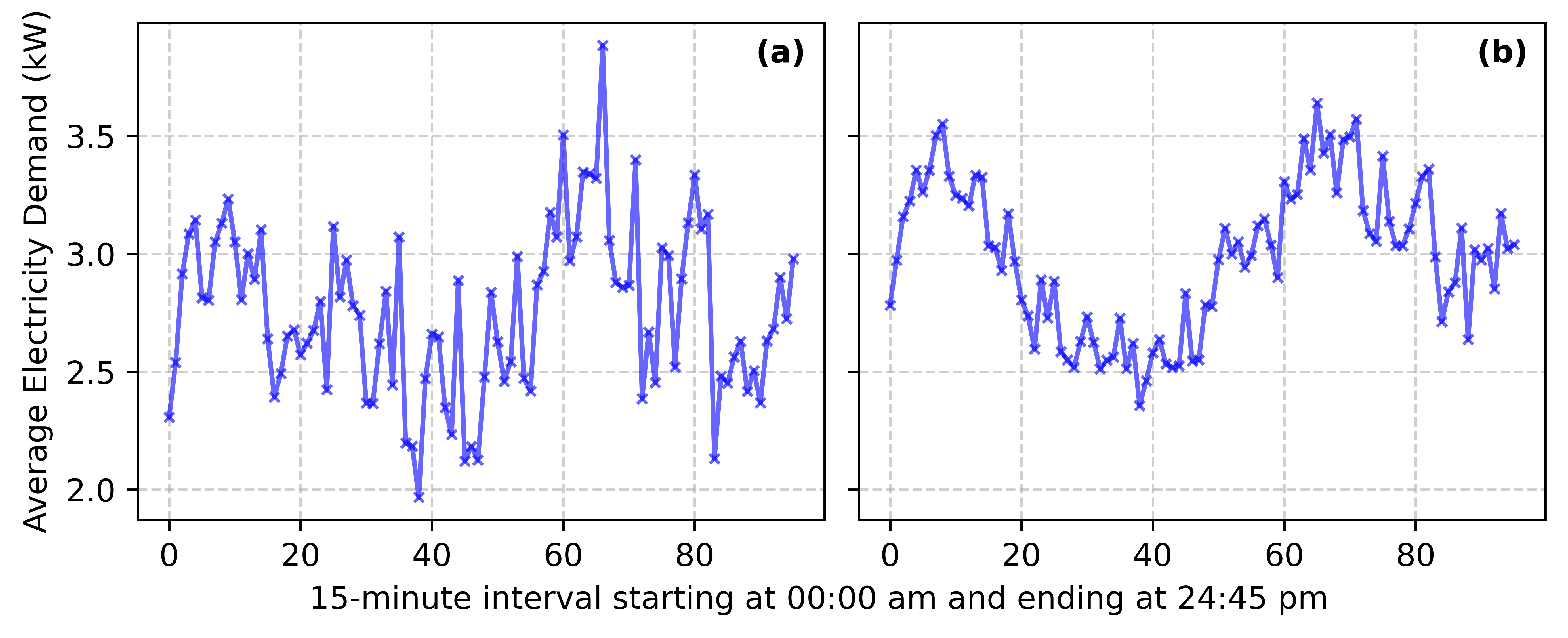}
  \caption{Plot of average demand aggregated over \(25\) houses \textbf{(a)} and \(150\) houses \textbf{(b)} over period of \(1\) day of \(15\)-minute intervals showing higher stochasticity in lower house aggregation.}
  \label{fig:low aggregation}
\end{figure}
The impact of low-aggregation illustrated in Figure \ref{fig:low aggregation} compares the average electricity demand of \(10\) houses against that of \(150\) houses. Lower aggregation leads to higher variability and uncertainty in demand, complicating accurate forecasting efforts. 

Demand forecasting of electricity can broadly range from short-term (days, hours, or real-time) to long-term (months or years) \cite{HONG2014357}. In this research, we are interested in the day-ahead short-term forecasting of aggregate demand, useful for unit commitment and economic dispatch planning in the energy markets. Broadly, two approaches have been used in the literature for demand forecasting: physics-based and statistical-based. Physics-based models often rely on system domain knowledge like the insulation characteristics of the house, HVAC system specifications, and residents' behavioral patterns to simulate the electricity usage over time \cite{SWAN20091819, ZHAO20123586}. On the other hand, statistical-based methods treat demand as time-series that can be learned from historical data. These models range from linear statistical models (e.g., ARIMA, Linear Regression (LR)) \cite{FUMO2015332, Kovacevic01042014} to more advanced ML methods (e.g., Neural Networks (NN), tree-based ensemble methods) \cite{syed2021deep, BEDI20191312, YILDIZ20171104, AzizEzzat11022025}. Statistical models capture correlation and patterns in the historical data without explicitly needing physical information of the systems.

However, because electricity demand is inherently noisy and stochastic, single-value forecasts often fail to fully reflect the spectrum of possible outcomes. Consequently, forecasting intervals -- which offer a range of potential demand scenarios -- have gained increasing prominence (e.g., \cite{Li2017ForecastingLoad}). Thus, in this research, our interest lies in determining the prediction interval for day-ahead electricity demands, under low-aggregation conditions. Consequently, we will apply ML-based approaches that depend on historical aggregate demand data. 

Tree-based ensemble models are among the most widely studied and effective Machine Learning (ML) methods for point forecast prediction, as evidenced by comprehensive surveys such as those by \cite{mienye2022survey} and \cite{yang2023survey}. These models leverage ensemble learning, a paradigm that combines multiple base learners to enhance predictive accuracy, a principle robustly demonstrated in works like \cite{bergmeir2016bagging}. Their success has led to diverse applications across energy systems, including wind and solar power generation forecasting (\cite{lee2020wind, voyant2018prediction, li2018interval}), short-term electricity demand prediction (\cite{yang2022interval, narajewski2020ensemble}), and building load estimation (\cite{wang2015review}). Ensemble methods broadly fall into two categories: boosting and bootstrapping. Boosting algorithms, such as Gradient Boosting Regression (GBR) and Light Gradient Boosting Machines (LGBM), iteratively train smaller trees to construct a strong predictive model by focusing on residual errors. In contrast, bootstrapping techniques like Random Forests (RF) generate parallel constituent trees trained on resampled subsets of the data, aggregating their outputs to reduce over-fitting and improve generalization.

Beyond point forecasts, ensemble methods also enable prediction interval estimation. For instance, \cite{JMLR:v7:meinshausen06a} showed that Random Forests, when adapted as Quantile Regression Forests, leverage predictions from constituent trees to model the full conditional distribution of outcomes. Similarly, gradient boosting frameworks like GBR and XGBoost can estimate uncertainty by replacing standard loss functions with quantile-specific objectives, training separate models for distinct quantiles (e.g., 0.05, 0.50, 0.95) and deriving intervals from the upper and lower bounds. These approaches are particularly valuable in energy forecasting, such as day-ahead electricity demand prediction, where quantifying uncertainty around short-term fluctuations is critical for risk-aware decision-making.

A straightforward limitation of training such ensemble-based point or prediction forecast method is the high computational time requirement. Training multiple models either in parallel or sequentially can be resource-intensive, demanding substantial processing power and memory. This issue is exacerbated when quantile-specific models (e.g., 0.05, 0.95) are trained separately. Furthermore, the residual errors of point forecasts may show non-stationary behavior due to sharp demand fluctuations and external factors (e.g. weather anomalies). For instance, electricity demand residuals may display skewed or multi-modal distributions, violating the stationary assumptions. Additionally, while ML models assume errors are independent and identically distributed (IID), the work in \cite{SANTOSJUNIOR2023119614} shows that the residuals from real-world time-series data inherently violate the IID assumption due to temporal dependencies (e.g., autocorrelation, seasonality). 

Instead of directly using the ensemble methods like RF, GBR, and LGBM, our work builds on the simplified approach of residual sampling proposed by \cite{Hyndman2021Forecasting:Practice}, where rather than training a full ensemble of models, historical residual errors from a single point forecast are directly bootstrapped and added to the present point forecasts. This method reduces computational complexity since only one model is trained for the point forecasts. The bootstrapping scheme used here and introduced by \cite{Efron1979BootstrapJackknife} assumes that the historical residuals would be homogeneous (\cite{Clements2007BootstrapSeries, pan2016bootstrap}) and would follow the same distribution as the future residuals. However, as noted earlier, we shall see in Section \ref{sec:problem definition} that the residuals obtained from point forecasts are heterogeneous and non-stationary. We shall see that the residuals of the electricity demand obtained by ML models have higher variance during the days of high electricity demand and have lower variance during low demand days. Additionally, the violation of the IID assumption of residuals due to the presence of autocorrelation is shown in Section \ref{sec:IE}. Thus, there is a presence of heteroscedasticity and temporal dependence of residuals obtained from point forecasts of ML models.

To overcome these limitations, we propose a Cluster-based Block Bootstrapping (CBB) algorithm. First, we employ a NN-based spectral clustering method (\cite{shaham2018}) to group days with similar demand patterns. This clustering process is designed to group together days whose residuals have approximately constant variance, thereby creating homogeneous sets of forecast errors. Once the clusters are formed, we implement a block bootstrapping technique in which contiguous segments of residuals (rather than individual, isolated errors) are sampled from the cluster that most closely matches the current forecast scenario. Block bootstrapping is crucial because it preserves the inherent temporal dependencies and autocorrelation present in the data, leading to more realistic and reliable interval estimates.

By integrating clustering with block bootstrapping, the CBB algorithm relaxes the strict IID and constant variance assumptions inherent in standard bootstrapping methods. This integration ensures that the prediction intervals are not only computationally efficient as only a single trained model is needed, but also robust enough to capture the complex variability of electricity demand. Our method is particularly well-suited for applications such as microgrid operations and energy market planning, where real-time decision-making depends on both rapid computation and accurate uncertainty quantification.

With that motivation, the contribution of this paper is as follows: 
\begin{enumerate}
    \item We train ML models to predict the point estimate of one-day-ahead electricity demands and develop a clustering algorithm to group similar days based on demand, organizing residuals accordingly.
    \item We design a bootstrapping scheme for constructing prediction intervals by sampling the residuals in blocks. This scheme selects the closest cluster based on the similarity between point forecasts and cluster centroids.
    \item To demonstrate the effectiveness of our approach, we compare the quality and accuracy of prediction intervals with tree based ensemble quantile methods and the state-of-the-art Prophet model. Results show improved prediction interval quality with out clustering scheme, achieving comparable accuracy. 
\end{enumerate}

The paper introduces electricity demand data in Section \ref{section:data}, outlines the problem definition, and presents point estimate results in Sections \ref{sec:problem definition} and \ref{sec:IE}. We propose the CBB algorithm in Section \ref{CBBA}.
Results of our algorithm are compared with other bootstrapping methods in Section \ref{sec: results}, demonstrating comparable performance with reduced computation time compared to baseline algorithms. Sections \ref{sec:future_work} and \ref{sec:conclusion} provide future research directions and concluding remarks, respectively.

\section{Data Collection} \label{section:data}
Training ML models requires large, high-resolution data to achieve accurate electricity load predictions. Projects like the Northwest Energy Efficiency Alliance's Residential End Use Load Research (EULR) and Pecan Street Austin have significantly advanced this domain by providing rich datasets with exceptional spatial and temporal granularity. For instance, the EULR initiative meticulously collected electricity demand data at one-minute intervals from homes spanning the Northwestern region, encompassing states such as Washington, Oregon, Montana, and Idaho. Moreover, these datasets incorporate crucial environmental factors like temperature, humidity, and atmospheric conditions, thereby enriching the training data available to ML models. Such comprehensive datasets empower ML models to achieve greater accuracy and robustness in predicting electric load behaviors.

We shall use data from the EULR project to train and evaluate the models for aggregate electricity demands. The EULR project is a regional study designed to gather accurate electricity demand profiles that could help us in understanding contemporary electricity end-use patterns. While the project collects data for every minute interval, it has provided public access to the 15-minute interval data of electricity demand in residential and commercial sites for research purposes. Since the inception of the project in 2020, data has been collected from around 400 sites. 
\begin{figure}[htpb]
    \centering
    \includegraphics[scale=1]{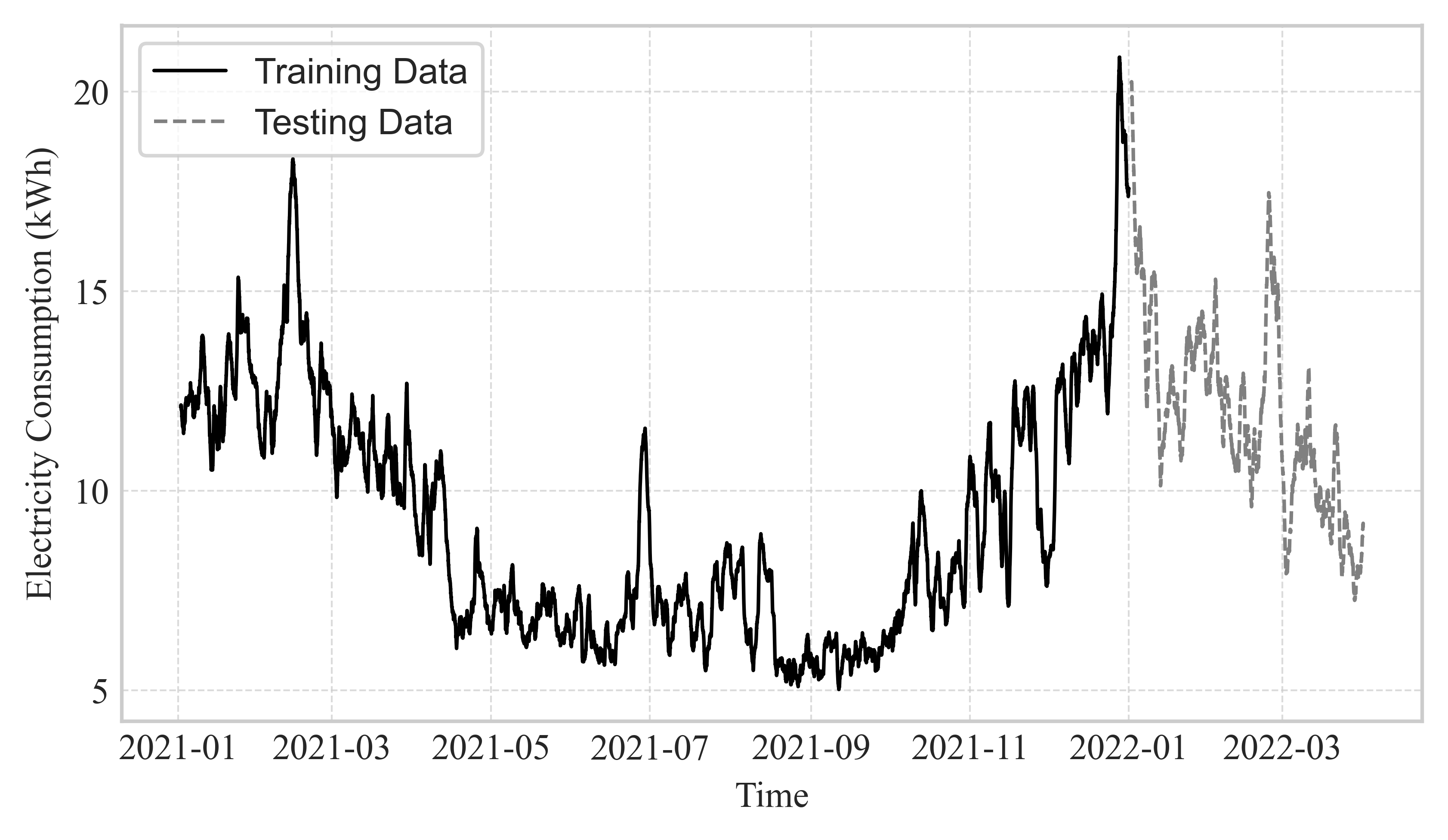}
    \caption{One-day moving average of aggregate electricity demand for \(50\) sites in Washington}
    \label{fig:moving average WA}
\end{figure}
The data provided in EULR consists of electricity drawn by the residential site's main supply line as well as at some of the major electrical appliances. The sites that have solar generation are removed as only the data on net electricity consumption is provided and, as a result, the time-series of electricity demand and solar generation cannot be separated for sites with solar generation. As a result, in this paper, we train our models on the electricity demand registered at the site's main supply line without any solar power generation. Compared to all the states mentioned earlier, the data for the highest number of residential sites were recorded in Washington state. The number of units from Washington for which data were continuously collected from the year 2020 to 2022 is \(50\). This is still considerably low and thus mimics a scenario where prediction for fewer households is needed as in a small Microgrid. Figure \ref{fig:moving average WA} shows the one-day moving average (96 intervals of 15 minutes) for the aggregate electricity demand of these \(50\) sites. The effects of annual seasonality can be seen as there is a downward trend in demand from the month of March to May and an upward trend from October to January. We describe the ML models in Section \ref{sec:problem definition} for which data from the year 2021 is used as a training sample and the data from the first quarter of 2022 is used for testing. The train-test split will remain the same in all of the following sections. We begin defining the problem setup and show the results of ML point estimates in the following section.

\section{Problem Definition} \label{sec:problem definition}
The objective of this study is to accurately forecast the prediction interval of the one-day-ahead aggregate electricity demand of the \(50\) residential sites. The interval prediction model in this research is based on residuals obtained from point estimates of the ML model's forecast. This section explains the inputs to the ML model and compares the results of the point estimates of the implemented ML models. Furthermore, Section \ref{sec:IE} formalizes the results of the point estimates discussed here and presents the necessary elements required for interval prediction.

Recall from Section \ref{section:data} that the data from the year 2021 is used as the training data. Each day in the training set is represented by $j$ where $j\in J=\{1,\  2,\ldots ,\ 365\}$. Further, the daily aggregated demand can be divided into 96 intervals represented by $i$ such that $i\in I=\{1,\  2\,\ldots,96\}$ with $i=1$ representing time $00:00:00$, sequentially increasing in intervals of 15 minutes until $23:45:00$. The training data for time-series can be considered as labeled data of the form $(\mathbf{X}_i^j, y_i^j)$, where $\mathbf{X}_i^j$ is the input vector comprising of the lags and exogenous variables and $y_i^j$ is the observed demand for the $i_{th}$ interval on a $j_{th}$ day. The input lag and exogenous variables for the ML model are selected as follows.

\subsection{Input Variable Selection}
\begin{figure}[h]
    \centering
    \includegraphics[scale=0.5]{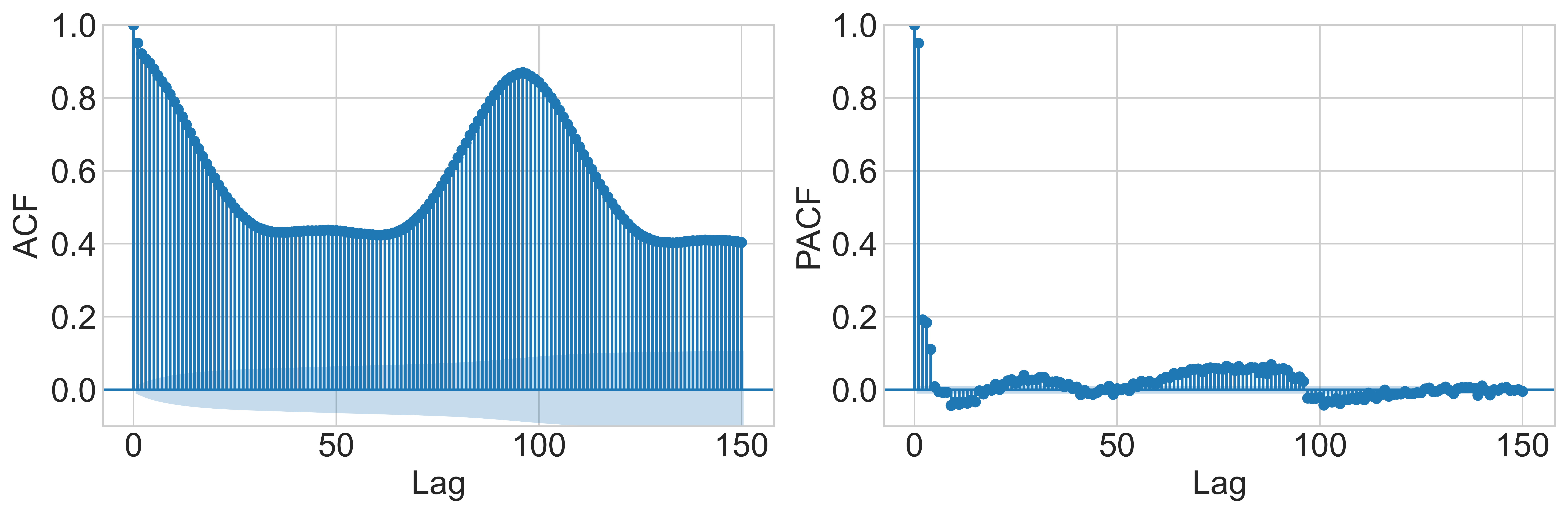}
    \caption{ACF plot (left) and PACF plot (right) of electricity demand}
    \label{fig:combinedcf}
\end{figure}

The plot of the Partial Auto-Correlation Function (PACF) is used by auto-regressive models to measure the correlation between the observed values of time-series (\cite{elsaraiti2021time}), in our case, the electricity demand of $y_i^j$ to $y_{i-k}^j$ for different values of $k$. The PACF for electricity demand data on the training set is plotted on the right-hand side of Figure \ref{fig:combinedcf} which shows the dependence of the demand $y_i^j$ on $y_{i-1}^j$ and $y_{i-2}^j$ values. It should be noted that since we are making a multi-horizon prediction for a one-day-ahead period, the lag values or the observed data during the $i-1$ and $i-2$, for $i>1$ would not be available for $i_{th}$ interval prediction. However, the Auto-Correlation Function (ACF) at the left-hand side of Figure \ref{fig:combinedcf} suggests that the electricity demand during the interval $i$ is correlated with the demand seen during the same interval of the previous day. Based on these observations from the PACF and ACF plots, the observed values $y_{i-1}^{j-1}$ and $y_{i-2}^{j-1}$ can serve as naive estimates for the two lag input variables for the prediction of demand in interval $i$.

\begin{figure}[h]
    \centering
    \includegraphics[scale=0.7]{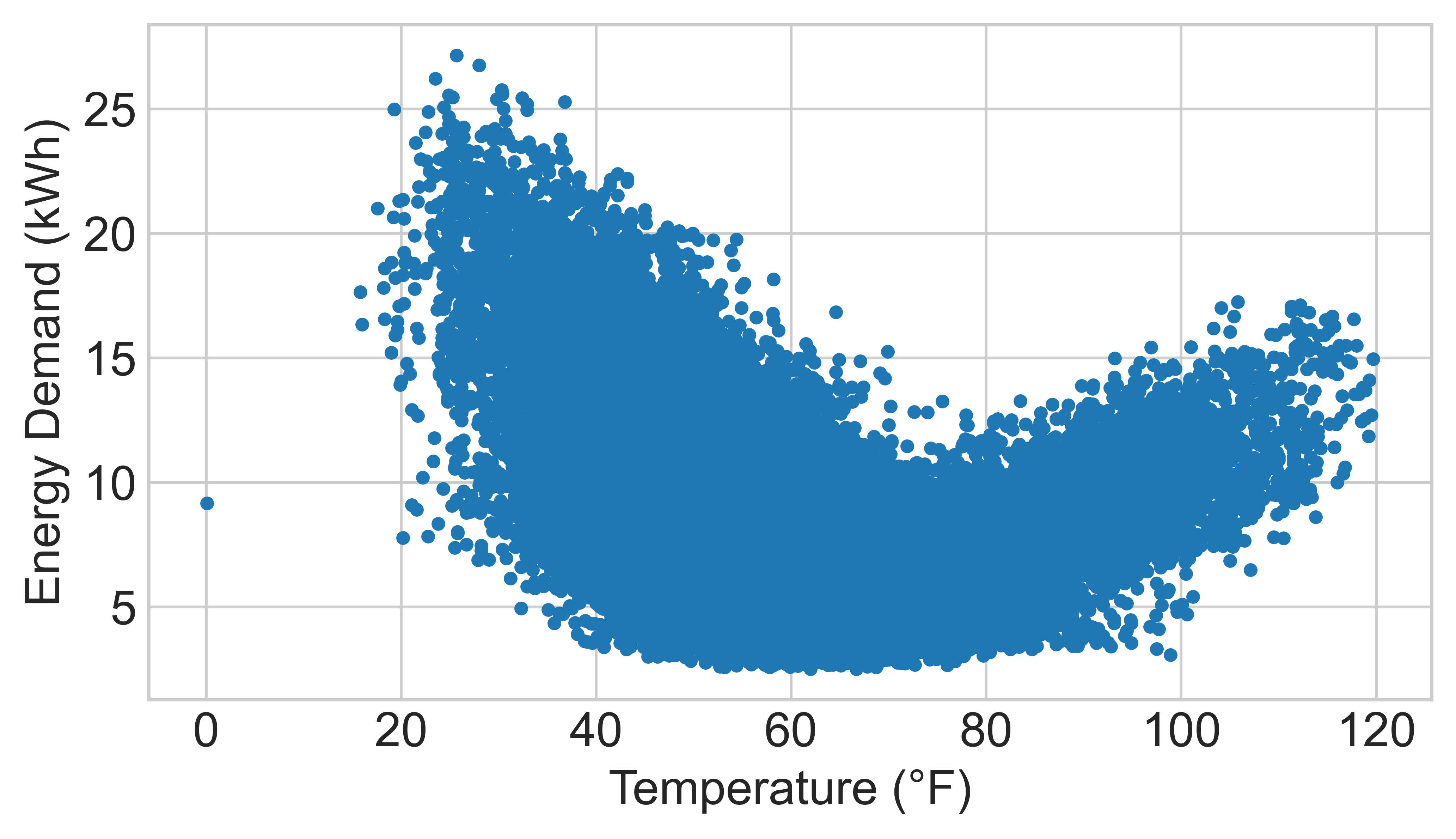}
    \caption{Temperature vs. Electricity Demand}
    \label{fig:demvstemp}
\end{figure}

We shall now look at the input exogenous variables used by the ML model. The calendar effects of a quarterly period of a year and holidays, including weekends and national holidays, are shown to affect electricity demand (\cite{Son2022Day-AheadProfiles}, \cite{U.S.Analysis}). Also, the dependence of the electricity demand on temperature is seen in Figure \ref{fig:demvstemp} where more electricity is required at lower temperatures, indicating the use of space heating units, and at higher temperatures as a result of using space cooling units in residential sites.
The temperature for any interval for a given day is the day-ahead predicted temperature from the nearest NOAA (National Oceanic and Atmospheric Administration) station. Thus, quarterly effects, holidays, and temperature predictions are considered as the input exogenous variables to the ML model.

Considering lag and exogenous variables, the input vector $\mathbf{X}_i^j$ for $j_{th}$ day and $i_{th}$ interval is thus defined as follows.

$$\mathbf{X}_i^j=(x_{i1}^j,\ x_{i2}^j, \ x_{i3}^j, \ x_{i4}^j,\ x_{i5}^j)$$
where,
\begin{align*}
x_{i1}^j & =y_{i-1}^{j-1} \quad \text{estimate for input lag variable of} \  y_{i-1}^{j},\\
x_{i2}^j & =y_{i-2}^{j-1} \quad \text{estimate for input lag variable of} \ y_{i-2}^{j},\\
x_{i3}^j & =\text{predicted temperature in Fahrenheit,}&\\
x_{i4}^j & =
  \begin{cases}
   0        & \text{Jan-Mar} \\
   1        & \text{Apr-Jun} \\
   2        & \text{Jul-Sep} \\
   3        & \text{Oct-Dec,} 
  \end{cases}
  \\
x_{i5}^j & = 
  \begin{cases}
   1        & \text{Holidays and Weekends (Saturday and Sunday)} \\
   0        & \text{other days.} 
  \end{cases}
\end{align*}

\begin{figure}[htb]
    \centering
    \includegraphics[scale=0.75]{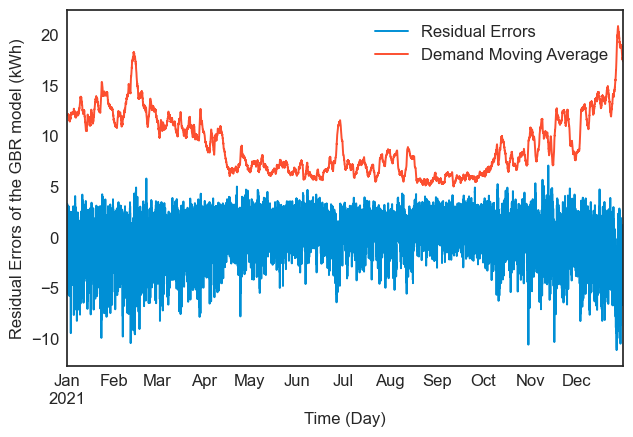}
    \caption{Residual errors of GBR on the training set with moving average of observed demand}
    \label{fig:res_err}
\end{figure}

We consider the ML model of the form $\hat{y}_i^t=\hat{f}(\mathbf{X}_i^j)$, where $\hat{f}$ is a real-valued function approximated by ML models. 
The usual assumption on the residual errors of such a model here denoted by $z_i^j=y_i^j-\hat{y}_i^j$ is that they are IID. 
As can be seen in Figure \ref{fig:res_err}, the residuals are centered around $0$ and the variance of the residuals is higher in the months of January to March, decreases until July, and again increases from August to December. This residual pattern follows the electricity demand with higher variance during the days of higher electricity demand and vice versa, representing non-stationarity.

\subsection{Point Estimate Metrics} \label{sec: pointestimatemetrics}
The point estimates on the testing set are generated by an expanding window technique on the training set. The current test day observations are added to the training set and a new training model is obtained for the next test day predictions. The moving window proceeds by first predicting the day-ahead demand of the test day $j'$ and then adding the label $\mathbf{X}_i^{j'}$ of the day to the training set for sliding window prediction, where $j'\in J'=\{1,\  2,\ldots,\ 90\}$ denotes the index of test days. 

The model errors of the training and testing data are shown in Table \ref{tab:pointforecast}. The absolute deviations from the observed demand are highest for GBR on test data compared to LR and LGBM. The lower error metrics on the LGBM model denote better point estimates on the test set.

ML models are susceptible to over-fitting on the training set resulting in the lower error on the training set and higher errors on the test set. If the training errors are directly bootstrapped for the interval estimation of the test day, the intervals would be narrow due to the over-fitting problem. We overcome this problem by replacing the errors of the training set with the errors on the test set sequentially, which is further described in Section \ref{CBBA}.

\begin{table}[htb]
\centering
\caption{Model performance for point forecasts}
\begin{tabular}{lcccccc}
\toprule
\textbf{Scores} & \textit{LR} Train & \textit{LR} Test & \textit{GBR} Train & \textit{GBR} Test & \textit{LGBM} Train & \textit{LGBM} Test \\
\midrule
\textbf{MAE}   & 1.325 & 1.659 & 1.087 & 1.473 & 1.080 & 1.474 \\
\textbf{MSE}   & 3.053 & 4.519 & 2.018 & 3.474 & 1.988 & 3.490 \\
\textbf{RMSE}  & 1.747 & 2.126 & 1.420 & 1.864 & 1.410 & 1.868 \\
\textbf{MAPE}  & 15.47\% & 14.91\% & 12.72\% & 13.40\% & 12.66\% & 13.36\% \\
\bottomrule
\end{tabular}
\label{tab:pointforecast}
\end{table}

\section{Interval Estimation} \label{sec:IE}
The proposed model for interval estimation of the electricity demand involves the use of residual errors obtained by the ML models seen in the previous section. We define and formalize the need for residual blocks and the memory clusters in this section.

\subsection{Residual Block} \label{sec: residual blocks}
We adopt a non-parametric approach to obtain the prediction intervals for electricity demand, where the residual errors are re-sampled in order to build the intervals. We begin by building up notation for the residual errors. The observed forecast error on the training data for the ML model is given as follows
\begin{flalign}\label{eq:1}
z_i^j&=y_i^j-\hat{y}_i^j, \quad \forall i\in{I}\ ,\ j\in{J}, 
\end{flalign}
where $y_i^j$ is the observed demand and $\hat{y}_i^j$ is the predicted demand by the ML models.
We define a memory set $E$ of residuals, such that the elements are a tuple of the $j_{th}$ day errors, thus for the training set we can define $E$ as 
\begin{equation} \label{E}
    E=\{(z_1^1,z_2^1, .... , z_{96}^1),\ldots,(z_1^j,z_2^j, .... , z_{96}^j),\ldots,(z_1^{365},z_2^{365}, .... , z_{96}^{365})\}.
\end{equation}
Then the residual errors for test data are given by {\(\hat{z}_i^{j'}\)}
\begin{flalign}
    z_i^{j'} &=y_i^{j'}-\hat{y}_i^{j'}, \quad \forall i\in{I}\ ,\ j'\in{J'}, \nonumber \\
     y_i^{j'} &=\hat{y}_i^{j'}+z_i^{j'}  \label{y_hat}.
\end{flalign} 
The prediction interval for $y_i^{j'}$ can be built by bootstrapping for $z_i^{j'}$ from the residual error set $E$, such that $y_i^{j'}=\hat{y}_i^{j'}+\hat{z}_i^{j'}$ if the errors are identically distributed. Thus, we shall first discuss the case of the traditional IID bootstrapping method. This method considers that the future errors of the test set are similar to the past errors so that $\hat{z}_i^{j'}$ can be approximated with the bootstrapped values of the residual errors from the training set $z_i^j$. Thus the residuals could be randomly selected with replacement from the memory set of the training residual errors $E$, $N$ times, where $N$ is a large valued integer. Suppose $N=1000$, and $(z^*_{(1)}, z^*_{(2)},\ldots,z^*_{(1000)})_i^{j'}$ is the ordered set of the bootstrapped residuals for day $j'$ and interval $i$ randomly selected from memory $E$ with replacement, then the $5th$ and the $95th$ percentile values of the prediction interval are represented by $z^*_{(50)}$ and $z^*_{(950)}$, respectively. 

\begin{figure}[h]
    \centering
    \includegraphics[scale=0.5]{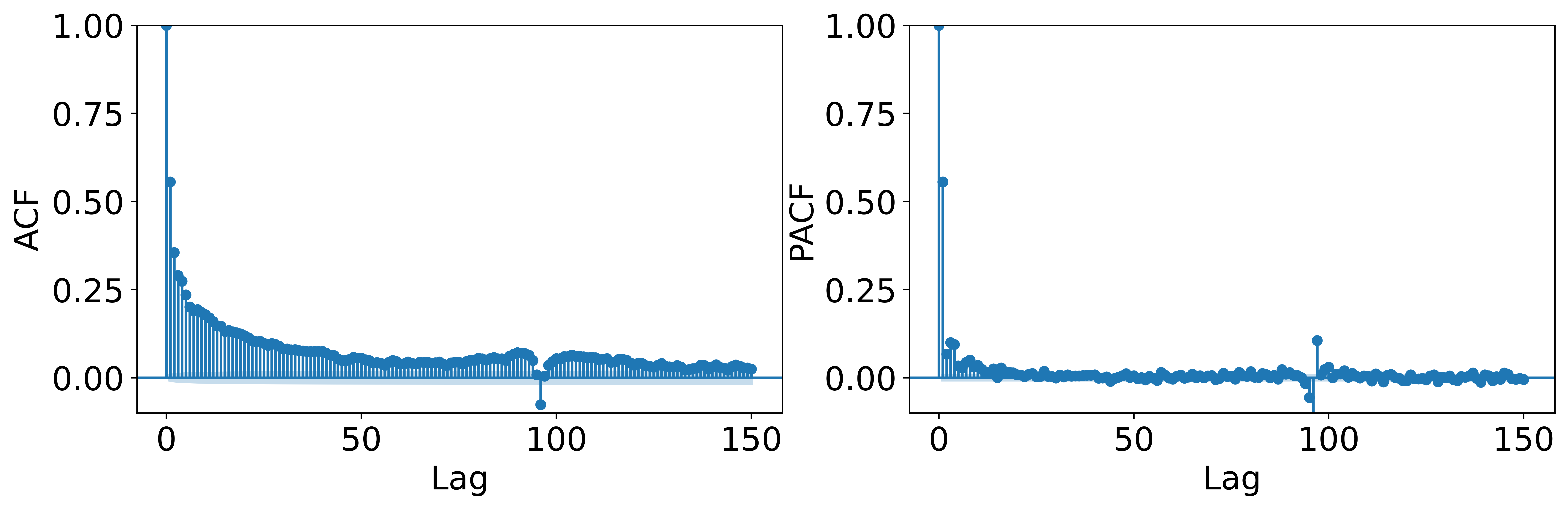}
    \caption{ACF plot (left) and PACF plot (right) of residuals of ML model}
    \label{fig:combinedcfres}
\end{figure}

However, the ACF and PACF plots of the residual series $\hat{z}_i^{j}$ presented in Figure \ref{fig:combinedcfres} indicate the existence of correlation among the residuals. As a result of this, the IID bootstrapping cannot be applied to the dependent data of residual electricity demand. Also, there are variations in the magnitude of the residuals on the training set as seen in Figure \ref{fig:res_err} indicating that the errors are not identical. This inadequacy of the IID bootstrapping method for dependent series is described in \cite{Singh1981OnBootstrap}. Instead of re-sampling a single observation of residuals at a time, non-overlapping contiguous blocks of residuals can be re-sampled; as a result, the structural dependence of the residuals can be preserved. Thus, the residual of the electricity demand isn't randomly selected from the memory $E$ and in order to account for the correlations among the errors, non-overlapping blocks of fixed length are drawn from the observed residual set and then joined. As predictions are made every $15$-minutes, a day is divided into $n=96$ intervals, which can be split into $b=16$ consecutive blocks of equal length $l=6$. We define the residual vector and the splitting rule as follows
\begin{flalign}
z^j&=(z_1^j,\ z_2^j,\ z_3^j,\ldots, z_{n}^j) \quad \forall \ j\in{J},\label{z_vec} \\
z^j&=(B_1^j,\ldots,B_b^j), \label{B_z}  \\
\text{such that}\quad B_k&=(z_{(k-1)l+1},\ldots,z_{kl}), \quad k=(1,\ .\ .\ . ,b), \nonumber
\end{flalign}
where the residual errors of the training data for $j_{th}$ day are given as a vector $z^j$, and the elements of this vector are calculated using Equation \eqref{eq:1}.

The accuracy of the bootstrapping in blocks is sensitive to the size of the blocks. As suggested in \cite{Politis2006AutomaticBootstrap}, the empirical block length of $n^{1/3}$ is used to select the block length $l$.

\subsection{Clustering Approach} \label{section: clustering}
The residuals used for constructing intervals should ideally be homogeneous. However, in the case of our ML model, as shown in Figure \ref{fig:res_err}, the residuals are influenced by the magnitude of the electricity demand. We thus first cluster on the similar days based on the electricity demand, and then store the residuals of these similar days in groups according to their demand clustering. The objective here is to form groups where the residuals within each group have similar magnitudes, but vary across groups.

While traditional unsupervised learning methods like k-means can be employed to cluster these residuals, spectral clustering algorithms are often more effective due to their ability to manage non-convex clusters and high-dimensional input. In this study, we propose an NN-based clustering approach. This method leverages a specialized loss function [\cite{shaham2018}] that embeds the input demand vector into a low-dimensional space and clusters these vectors based on a similarity function applied to the input vectors.
This NN-based clustering scheme aims to enhance clustering performance by capturing complex nonlinear relationships in the data, offering a more robust solution compared to conventional clustering methods.

\begin{figure}[h!]
    \centering
    \includegraphics[width=0.7\linewidth]{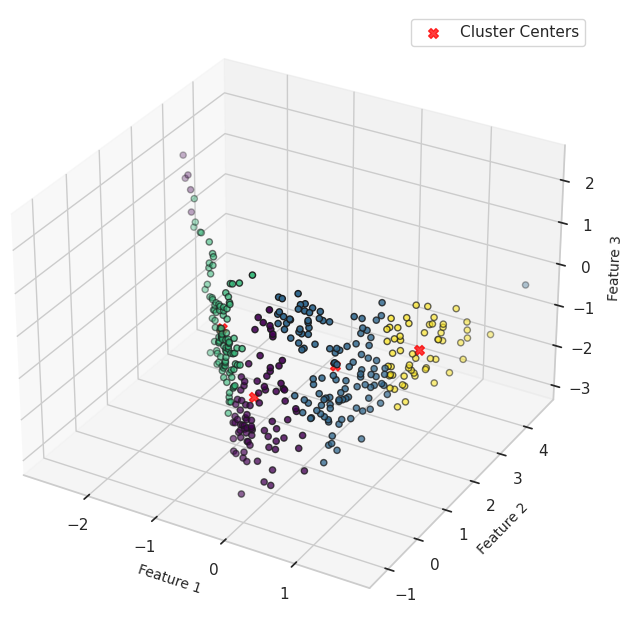}
    
    \caption{Clusters of days based on electricity demand are shown in the embedded space, with clustering performed using the first three dimensions of the embedded feature space. The corresponding cluster centers are also indicated.}
    \label{fig:embed cluster}
\end{figure}

The similarity function, $w(y^j, y^r)$ where, $w:\mathbb{R}^n\times\mathbb{R}^n\rightarrow[0, \infty)$ and $j,r\in J$ calculates the pairwise symmetric euclidean distance between demand vector $y^j$ and $y^r$. Given such a function, the goal is to embed the similar input vectors in the embedding space using NN with a loss function as follows,

\begin{equation}\label{eq:custom loss}
    L_{clustering}(\theta)=\mathbb{E}[w(y^j,y^r)]\|l^j-l^r\|^2 
\end{equation}

where, $l^j, l^r  \in \mathbb{R}^{n'}$ are the outputs of NN such that $F_\theta:\mathbb{R}^n\rightarrow\mathbb{R}^{n'}$ and $\theta$ are NN's parameters. From  Equation \eqref{eq:custom loss}, it is clear that the loss function is minimized when the embedding vectors $(l^k,l^j)$ are close to each other for high euclidean similarity $w(y^j,y^r)$. These embedding vectors are grouped together into four clusters, as shown in Figure \ref{fig:embed cluster}. Finally, the k-means clustering algorithm is applied to the embedding space just to label the already grouped clusters.

\begin{figure}[h!]
    \centering
    \includegraphics[width=0.85\linewidth]{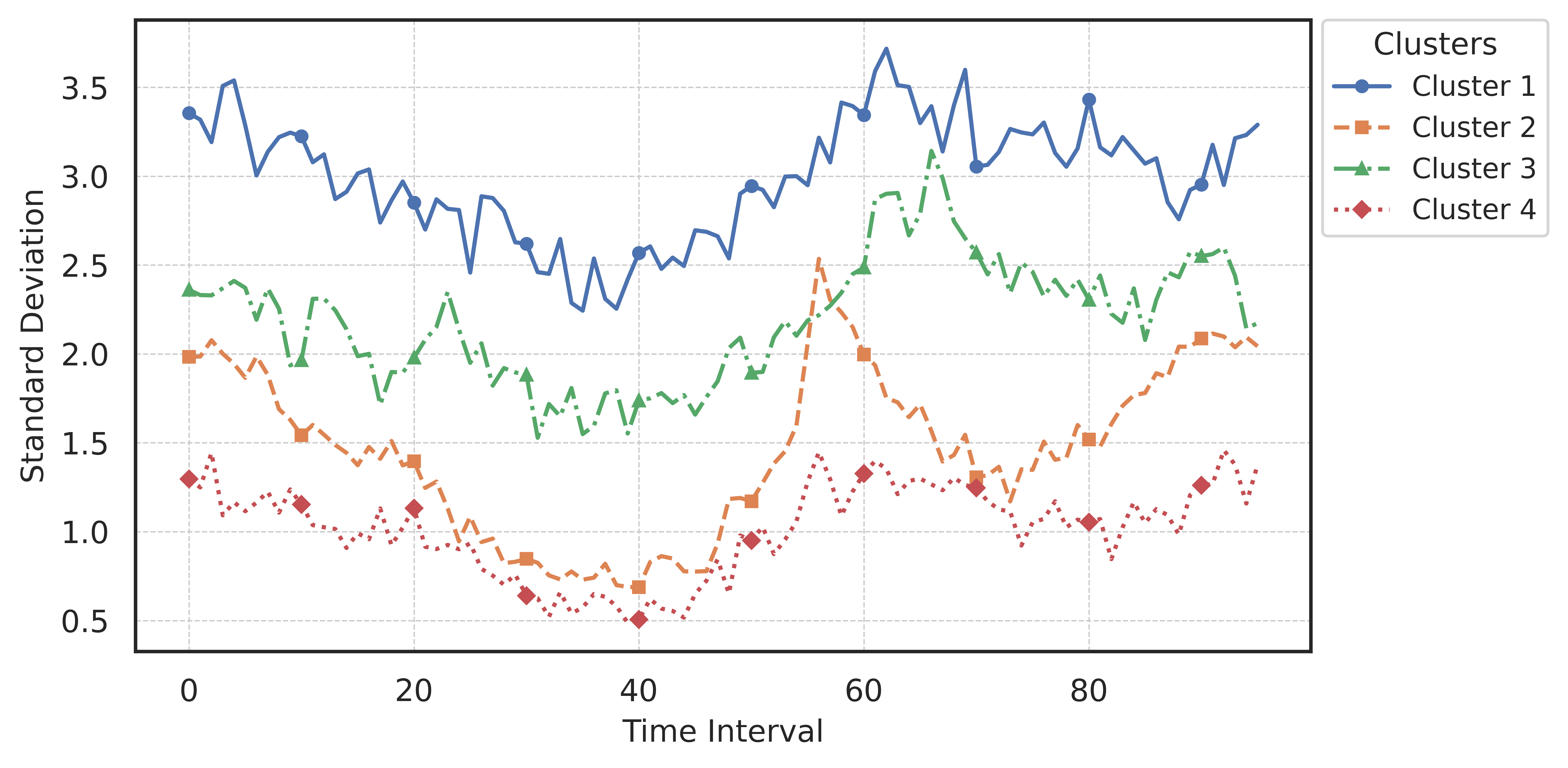}
    \caption{Standard Deviation of Residuals across Time Intervals for Different Clusters.}
    \label{fig:stndev clusters}
\end{figure}

In this experiment, the k-means clustering algorithm labels the $N_c=4$ clusters of embedded vectors $l^j$. It returns a set of centroids $l_{C_k}$, one for each of the $N_c$ clusters, with each embedded vector labeled by the centroid index $C_k,\ \forall k \in (1,\ldots, N_C)$.
The residuals are also grouped based on these clusters, reflecting the similarity in electricity demand. The standard deviation of the residuals within each group, shown in Figure \ref{fig:stndev clusters}, indicates distinct magnitudes of residuals across different demand clusters.

\subsection{Performance Metrics}

Our interest is in finding the quantile values during the time interval $i$ within which the values of electricity demand might lie with a probability $100(1-\alpha)\%$ which is the size of the prediction interval where $\alpha$ is the confidence level, $0\leq\alpha\leq1$. The predicted upper quantile and the lower quantile is denoted by $u_{\alpha,i}^j$ and $l_{\alpha,i}^j$ respectively for the time interval $i$ on day $j$.
 
The accuracy of the prediction interval model is measured by the number of times the prediction interval includes the true value. This proportion, known as \textit{Coverage Probability} ($CP(\alpha)$), for a confidence level $\alpha$, quantifies the reliability of the prediction interval $[l_{\alpha,i}^j,u_{\alpha,i}^j]$ and is defined,
 \begin{align}\label{eq:CP}
    CP(\alpha)=\frac{1}{|J|}\sum_{j\in J} \sum_{i\in I} \frac{\mathds{1}_{[l_{\alpha,i}^j\leq y_i^j\leq u_{\alpha,i}^j]}}{|I|},
\end{align}

\begin{align*}
    \text{where},\quad \mathds{1}_{[l_{\alpha,i}^j\leq y_i^j\leq u_{\alpha,i}^j]}&=1  \quad when\quad {[l_{\alpha,i}\leq y^j_i\leq u_{\alpha,i}]},\\
    &=0 \quad  otherwise.
\end{align*}

It should be noted that Equation \eqref{eq:CP} alone can be misleading because very wide intervals can trivially achieve high coverage by encompassing a broad range of values, which may not be practically useful. In order to measure the width and quality of the prediction intervals $[l_{\alpha,i}^j,u_{\alpha,i}^j]$, we use \textit{Winkler Score} $(WS(\alpha))$, proposed by \cite{Winkler1972AEstimation}, defined as follows,

\begin{align}\label{eq WS}
    WS(\alpha)_i^j=
    \begin{cases}
        u_{\alpha,i}^{j}-l_{\alpha,i}^{j}+\frac{1}{\alpha}(l_{\alpha,i}^{j}-y_i^{j})\quad &if\quad y^{j}_i<l_{\alpha,i}^{j},\\
        u_{\alpha,i}^{j}-l_{\alpha,i}^{j} \quad  &if\quad l_{\alpha,i}^{j}\leq y^{j}_i\leq u_{\alpha,i}^{j},\\
        u_{\alpha,i}^{j}-l_{\alpha,i}^{j}+\frac{1}{\alpha}(y^{j}_i-u_{\alpha,i}^{j})\quad &if\quad y^{j}_i>u_{\alpha,i}^{j},
    \end{cases}
\end{align}
For each interval $i$ on day $j$ the average metric is taken as follows:
\begin{align*}
    WS(\alpha)=\frac{1}{|J|}\sum_{j\in J} \sum_{i\in I} \frac{WS_(\alpha,i)^j}{|I|}.
\end{align*}

It can be seen that Equation \eqref{eq WS} penalizes intervals that either miss the true value or are excessively wide, providing a more comprehensive measure of the prediction interval's quality. This ensures that the intervals are not only reliable in terms of coverage but also useful in practice by being sufficiently narrow.

\section{Cluster-based Block Bootstrap Algorithm} \label{CBBA}

In the Section \ref{sec:IE}, we saw the methods to bootstrap residual blocks and to create clusters of similar days. In this section, we will combine these two methods together to generate the prediction intervals for one-day-ahead forecasts. The first step as shown in Algorithm~\ref{alg::cbb1} is to train the ML model $\hat{f}$ using the training data $(\mathbf{X}_i^j,y_i^j)$ and get the residual errors $z^j_i$ on the training set $j$. These training errors are stored in the memory set $E$ defined in Equation \eqref{E}. 

{\setstretch{1.0}
\begin{algorithm}
\caption{CBB Algorithm -- Training and Clustering}
\label{alg::cbb1}
\begin{algorithmic}[1]
\Require Historical training data $\{(\mathbf{X}_i^j, y_i^j)\}_{j=1}^{|J|}$
\Require Number of intervals per day $n$, block length $l$, clusters $N_c$
\Require Embedding function $F_\theta$

\State Train forecast model $\hat{f}$ on training data

\For{each day $j = 1$ to $|J|$}
    \For{each interval $i = 1$ to $n$}
        \State Compute residual $z_i^j = y_i^j - \hat{f}(\mathbf{X}_i^j)$
    \EndFor
    \State Form vector $z^j = (z_1^j, \dots, z_n^j)$, divide into $b = n/l$ blocks
\EndFor

\For{each day $j = 1$ to $|J|$}
    \State $y^j = (y_1^j, \dots, y_n^j)$
    \State Compute embedding $l^j = F_\theta(y^j)$
\EndFor

\State Cluster $\{l^j\}$ into $N_c$ groups using spectral clustering
\For{each cluster $C_k$}
    \State Compute centroid $l_{C_k}$ and store residuals $M_k = \{z^j : j \in C_k\}$
\EndFor

\Ensure $\hat{f}$, memory sets $\{M_k\}$, centroids $\{l_{C_k}\}$
\end{algorithmic}
\end{algorithm}
}

In the next step, we form clusters of indices representing similar days using an NN-based spectral clustering algorithm on the electricity demand $y^j$ for $j\in J$. The label of each cluster represented by $C_k$ has a centroid at $\bar{y}_{C_k}$ where $k\in \{1,\dots,N_C\}$. We represent the index of days clustered together in the $k_{th}$ cluster as $\{(1),\ldots, (|C_k|)\}$, where $|C_k|$ denotes the size of the $k_{th}$ cluster and $\{(1),\ldots,(|C_k|)\}$ are the clustered training data days partitioned off training set labeled $I$ such that $\{(1),\ldots,(|C_k|)\} \in I$.

The memory set of residuals $E$ is then partitioned to form the cluster memory set $M_k$, where $M_k$ is selected according to the days indexed in cluster $C_k$. Thus for every cluster label $C_k\in \{(1),\ldots ,(|C_k|)\}$ we get  $M_k=\{z^{(1)},\ldots ,z^{(|C_k|)} \}$. Using Equation (\ref{z_vec}) the set $M_k$ can be denoted in terms of residual blocks $$M_k =\{(B_1^{(1)},\ldots , B_b^{(1)}),\ (B_1^{(2)},\ \ldots, B_b^{(2)}),\ldots,\ (B_1^{(|C_k|)},\ldots,B_b^{(|C_k|)})\}$$ 
where number of blocks, $b=16$, and length of residual vector $n=96$ such that $n=b\times l$  as defined in Section \ref{sec: residual blocks}.

The model $\hat{f}$, the clustered residual sets $M_k$ and the centroid of the clusters $l_{C_k}$ for $k\in(1,\ldots, N_c)$ are now ready to evaluate the point estimates and construct the prediction intervals. As shown in Algorithm~\ref{alg:CBB2} the ML model $\hat{f}$, is used to get the point estimates $\hat{y}^{j'}$ for test period $j'$ for $j'\in J'$. The point estimates of the test day $\hat{y}^{j'}$ are used as an input to the forward pass of $F_\theta$ and the output embedding vector $\hat{l}^{j'}$ of test day is obtained. The closeness of test day embedded vector $\hat{l}^{j'}$, is evaluated with every cluster's centroid $l_{C_k}$ using euclidean distance and the closest $k_{th}$ residual cluster memory $M_k$ is selected to bootstrap the block residuals for the $j'th$ test day.

The test day is also divided into $16$ non-overlapping blocks of size $6$, and for the $i_{th}$  interval block of the test day, we bootstrap $N=1000$ times, residual blocks $B_i^{(n)}$ from the selected cluster $M_k$ randomizing on $n$ such that $n\in(1,\ldots,|C_k|)$ and repeat this process for each $i\in(1,\ldots,16)$. Then for the $i_{th}$ time interval block we can get $N$ bootstrap residual block samples and build  $\textit{\textbf{B}}_i=(B_1^*,\ldots ,B_N^*)_i$\footnote{The star notation on x* indicates that x* isn't the real data set but the randomized, re-sampled or bootstrapped version of x}. The sets $\textit{\textbf{B}}_1,\ldots,\textit{\textbf{B}}_{16}$ are then joined sequentially to form the prediction interval for the test day.

{\setstretch{1.0}
\begin{algorithm}[htbp]
\caption{CBB Algorithm -- Testing, Cluster Selection, and Residual Sampling}\label{alg:CBB2}
\begin{algorithmic}[1]
\Require Test day inputs $\{\mathbf{X}_i^{j'}\}_{i=1}^n$
\Require Trained model $\hat{f}$, embedding function $F_\theta$
\Require Cluster centroids $\{l_{C_k}\}$, residual sets $\{M_k\}$, number of samples $N$

\State Predict $\hat{y}_i^{j'} = \hat{f}(\mathbf{X}_i^{j'})$ for $i = 1, \dots, n$
\State Form vector $\hat{y}^{j'} = (\hat{y}_1^{j'}, \dots, \hat{y}_n^{j'})$
\State Compute embedding $\hat{l}^{j'} = F_\theta(\hat{y}^{j'})$

\State Identify nearest cluster:
\[
k^* = \arg\min_k \|\hat{l}^{j'} - l_{C_k}\|
\]
\State Retrieve residual memory $M_{k^*}$

\State Partition the day into $b = n/l$ blocks

\For{each block $r = 1$ to $b$}
    \State Sample $N$ residual blocks $B_r^*(1), \dots, B_r^*(N)$ from $M_{k^*}$ with replacement
    \State Add sampled blocks to forecast block $\hat{y}^{j'}_{\text{Block}_r}$ to create bootstrap trajectories
\EndFor

\State Concatenate sampled blocks to form $N$ full-day bootstrap demand vectors $\{y^{*(1)}, \dots, y^{*(N)}\}$

\For{each interval $i = 1$ to $n$}
    \State Extract $\{y_i^{*(1)}, \dots, y_i^{*(N)}\}$
    \State Compute quantiles: $l_{\alpha,i}^{j'}$ and $u_{\alpha,i}^{j'}$
\EndFor

\Ensure Prediction intervals $[l_{\alpha,i}^{j'}, u_{\alpha,i}^{j'}]$ for all $i$

\end{algorithmic}
\end{algorithm}
}
Due to the over-fitting issues identified in Section~\ref{sec: pointestimatemetrics}, bootstrapping residuals solely from the training set yields excessively narrow prediction intervals. To address this limitation, we propose an adaptive clustering that dynamically updates the residual memory. Rather than relying on static training-set residuals stored in $M_k$, we progressively replace them with the model's test-set residuals as new observations become available. Specifically, for each observed test day $j'$, we:
\begin{enumerate}
    \item Identify its corresponding training day $j$ (where $j' \equiv j$),
    \item Update the residual $z^j$ with the test error $z^{j'}$, 
\end{enumerate}
This iterative refinement ensures that bootstrapped residuals reflect the model's real-time performance, improving interval estimation accuracy.

\section{Results} \label{sec: results}
In this section, we show the results of the CBB algorithm and compare it against the tree-base ensemble quantile models along with the Prophet. In order to show the improvement in quality of the prediction intervals due to clustering, we will also compare the CBB results with residual bootstrapping without clustering. For the residual bootstrapping models LR, GBR, and LGBM are used for point estimation. To make a direct comparison with residual bootstrapping, the quantile regression models also use LR, GBR, and LGBM as weak learners.

For ease of notation, we will use $WS$ and $CP$ instead of $WS(\alpha)$ and $CP(\alpha)$ respectively in the following sections. The \( WS \) metric penalizes wider intervals and wrong predictions, thus a smaller value of \( WS \) is desirable. Conversely, \( CP \), measures the accuracy of the prediction intervals in capturing the true parameter value. Together, \( WS \) represents the quality of the prediction intervals by penalizing deviations and excessive width, while \( CP \) ensures their accuracy by quantifying how often the true value falls within the intervals.

\subsection{Prophet}
Prophet is a time-series forecasting method that uses an additive model to accommodate non-linear trends, incorporating yearly, weekly, and daily seasonal patterns, as well as holiday influences. Prophet takes as input the date-time features of electricity demand along with temperatures as exogenous variables. The results of Prophet on the electricity demand prediction interval of test set is shown in the Table \ref{tab:Prophet}. For the Prophet model, separate models must be trained for each different confidence level $\alpha$.

\begin{table}[h!]
\centering
\caption{Prediction interval performance of \textit{Prophet} model at various confidence levels}
\begin{tabular}{llccccc}
\toprule
\textbf{Model} & \textbf{Metrics} & \textbf{85\%} & \textbf{90\%} & \textbf{95\%} & \textbf{99\%} & \textbf{Train time (sec)} \\
\midrule
\multirow{2}{*}{\textit{Prophet}} 
 & \textit{WS} & 18.02 & 18.97 & 22.54 & 37.94 & \multirow{2}{*}{110.56} \\
 & \textit{CP} & 0.573 & 0.747 & 0.931 & 0.986 & \\
\bottomrule
\end{tabular}
\label{tab:Prophet}
\end{table}

\subsection{Quantile-Regression Benchmarks}
\label{sec:quantile_benchmarks}
To establish a reference point for the proposed CBB procedure, we estimate an entire conditional-quantile function. 
Let \(\mathcal Q=\{\alpha_{1},\ldots,\alpha_{K}\}\subset(0,1)\) be the set of
quantile levels of interest.  For each \(\alpha\in\mathcal Q\) we estimate an
independent model
\(\,f_{\alpha}:\mathbb R^{p}\rightarrow\mathbb R\)
that approximates the conditional \(\alpha\)-quantile of the response
variable~\(y\) given the predictor vector \(\mathbf X\).

\paragraph{Loss function.}
For a fixed \(\alpha\) the model parameters
\(\theta_{\alpha}\) are obtained by minimising the
\emph{pinball} (check) loss:
\begin{equation}
  \label{eq:pinball}
  \widehat{\theta}_{\alpha}
  \;=\;
  \arg\min_{\theta}
  \sum_{i=1}^{n}
      \rho_{\alpha}\!\bigl(
          y_{i}-f_{\alpha}(\mathbf X_{i};\theta)
      \bigr),
  \qquad
  \rho_{\alpha}(u)=u\bigl(\alpha-\mathbf 1_{\{u<0\}}\bigr).
\end{equation}
Solving~\eqref{eq:pinball} guarantees that, in expectation,
\(\mathbb P\!\bigl\{
  y \le f_{\alpha}(\mathbf X;\,\widehat{\theta}_{\alpha})
\bigr\} = \alpha\).

\paragraph{Prediction.}
Given a new covariate vector \(\mathbf x_{\text{new}}\), the estimated
conditional \(\alpha\)-quantile is
\[
  \widehat Q_{\alpha}(\mathbf x_{\text{new}})
  \;=\;
  f_{\alpha}\bigl(\mathbf x_{\text{new}};\,\widehat{\theta}_{\alpha}\bigr).
\]
Collecting the estimates
\(\bigl\{\widehat Q_{\alpha}(\mathbf x_{\text{new}})\bigr\}_{\alpha\in\mathcal Q}\)
yields an empirical conditional-quantile function—that is, a full predictive
distribution for the target variable at \(\mathbf x_{\text{new}}\).


\begin{table}[h!]
\centering
\caption{Prediction interval performance of ensemble models using quantile-based estimation across different confidence levels}
\begin{tabular}{llccccc}
\toprule
\textbf{Model} & \textbf{Metrics} & \textbf{85\%} & \textbf{90\%} & \textbf{95\%} & \textbf{99\%} & \textbf{Train time (sec)} \\
\midrule
\multirow{2}{*}{\textit{LR}} 
 & \textit{WS} & 8.462 & 9.205 & 10.506 & 13.400 & \multirow{2}{*}{327.97} \\
 & \textit{CP} & 0.742 & 0.827 & 0.905  & 0.975  & \\
\midrule
\multirow{2}{*}{\textit{GBR}}  
 & \textit{WS} & 7.419 & 8.041 & 9.133  & 12.547 & \multirow{2}{*}{199.99} \\
 & \textit{CP} & 0.743 & 0.830 & 0.916  & 0.975  & \\
\midrule
\multirow{2}{*}{\textit{LGBM}} 
 & \textit{WS} & 7.704 & 8.432 & 10.052 & 15.497 & \multirow{2}{*}{161.02} \\
 & \textit{CP} & 0.684 & 0.782 & 0.869  & 0.950  & \\
\bottomrule
\end{tabular}
\label{tab:quantile_ensemble_results}
\end{table}

Table~\ref{tab:quantile_ensemble_results} compares three bootstrap-aggregated
quantile ensembles on the test set.  The main findings are:

\begin{enumerate}
  \item \textbf{GBR delivers the highest-quality intervals.}  
        Across all nominal coverages (85\,–\,99\%), the
        gradient-boosted ensemble attains the lowest Winkler Score 
        (\(WS\)) \emph{and} the highest or joint-highest coverage probability
        (\(CP\)), indicating the best trade-off between sharpness and
        reliability.
        
  \item \textbf{LGBM is a fast, but slightly less reliable alternative.}
        LightGBM offers the second-sharpest intervals and
        the shortest training time (161 s, 
        \(\approx\!20\,\%\) faster than GBR and
        \(\approx\!50\,\%\) faster than LR).  However, its \(CP\)
        falls noticeably below the target at the narrower confidence levels,
        so practitioners must weigh speed against interval accuracy.

  \item \textbf{Linear Quantile Regression (LR) trails on sharpness and speed.}
        The linear model produces the widest intervals
        (largest \(WS\)).
        Its coverage matches GBR only at the 99\,\% level, providing limited benefit unless interpretability is paramount.

  \item \textbf{Computational cost remains a practical concern.}
        Because each bootstrap replicate demands a separate model fit, all ensemble approaches incur substantially higher run times than a single, non-aggregated model.  GBR offers the best interval quality per second of compute, while LGBM minimises wall-clock time when a slight drop in coverage is acceptable.
\end{enumerate}

\subsection{Block Bootstrap}
The results of the residual block bootstrapping without clustering are discussed in this section. We will simply refer to it as the Block Bootstrap algorithm. The prediction intervals are constructed similarly to CBB algorithm, i.e., by bootstrapping the non-overlapping residuals block but without clustering the similar days. This will help us understand the effect of clustering on quality of prediction interval as compared to the CBB algorithm.

\begin{table}[h!]
\centering
\caption{Prediction interval performance of ML point models using block bootstrap at various confidence levels}
\begin{tabular}{llccccc}
\toprule
\textbf{Model} & \textbf{Metrics} & \textbf{85\%} & \textbf{90\%} & \textbf{95\%} & \textbf{99\%} & \textbf{Train time (sec)} \\
\midrule
\multirow{2}{*}{\textit{Ridge}}   
 & \textit{WS} & 8.676 & 9.699 & 11.493 & 41.881 & \multirow{2}{*}{1.18} \\
 & \textit{CP} & 0.738 & 0.806 & 0.886  & 0.981  & \\
\midrule
\multirow{2}{*}{\textit{GBR}}  
 & \textit{WS} & 7.705 & 8.699 & 10.440 & 52.423 & \multirow{2}{*}{14.06} \\
 & \textit{CP} & 0.709 & 0.781 & 0.859  & 0.974  & \\
\midrule
\multirow{2}{*}{\textit{LGBM}} 
 & \textit{WS} & 7.735 & 8.722 & 10.459 & 48.800 & \multirow{2}{*}{29.23} \\
 & \textit{CP} & 0.719 & 0.784 & 0.866  & 0.976  & \\
\bottomrule
\end{tabular}
\label{tab:nocluster}
\end{table}

The performance of the Block Bootstrap is shown in Table \ref{tab:nocluster}. We observe the following:

\begin{enumerate}
    \item The GBR model achieves better performance due to lower \( WS \) and higher \( CP \) values compared to the LR and LGBM models, except for the \( WS \) value at the 85\% confidence interval size.
    \item The block bootstrapping approach demonstrates a significant reduction in computation time as compared to the ensemble models since only a point estimate model is trained.
    \item However, the quality of the prediction interval (\( WS \)) for block bootstrap model degrades compared to the ensemble models. We will see how clustering improves the quality while retaining the accuracy.
\end{enumerate}

\subsection{CBB algorithm}
The CBB algorithm proposed in Section \ref{CBBA} takes advantage of the clustering scheme and improves the quality of the residual block bootstrapping. In this section, we discuss the performance of the CBB algorithm, for which the results are shown in Table \ref{tab:CBBA}. 

\begin{table}[h!]
\centering
\caption{Prediction interval performance of ML point models using the CBB algorithm at different confidence levels}
\begin{tabular}{llccccc}
\toprule
\textbf{Model} & \textbf{Metrics} & \textbf{85\%} & \textbf{90\%} & \textbf{95\%} & \textbf{99\%} & \textbf{Train time (sec)} \\
\midrule
\multirow{2}{*}{\textit{LR}}   
 & \textit{WS} & 8.114 & 8.841 & 10.010 & 10.944 & \multirow{2}{*}{5.347} \\
 & \textit{CP} & 0.811 & 0.869 & 0.924  & 0.976  & \\
\midrule
\multirow{2}{*}{\textit{GBR}}  
 & \textit{WS} & 6.960 & 7.643 & 8.579  & 9.067  & \multirow{2}{*}{20.299} \\
 & \textit{CP} & 0.810 & 0.867 & 0.926  & 0.983  & \\
\midrule
\multirow{2}{*}{\textit{LGBM}} 
 & \textit{WS} & 8.059 & 8.930 & 10.345 & 12.500 & \multirow{2}{*}{33.023} \\
 & \textit{CP} & 0.790 & 0.833 & 0.910  & 0.976  & \\
\bottomrule
\end{tabular}

\label{tab:CBBA}
\end{table}

The results of CBB are summarized as follows:
\begin{enumerate}
    \item Similar to Block Bootstrapping, the quality and accuracy of the CBB algorithm with GBR point estimate model is the best due to its low \(WS\) and high \(CP\) values.
    \item The CBB algorithm shows a reduction in computation time as compared to the quantile regression method with only a slight increase compared to Block Bootstrapping due to time taken by clustering.
    \item The quality of the CBB algorithm \(WS\), greatly benefits due to the clustering scheme as the residuals are sampled from homogeneous groups. For instance, the prediction interval for a day with low point estimate demand is sampled using a residual cluster characterized by lower magnitude but constant variance. This approach prevents the sampling of high-magnitude residuals from days with high demand, thereby avoiding wider intervals in the prediction.
\end{enumerate}

\begin{figure}[h!]
    \centering
    \includegraphics[scale=0.65]{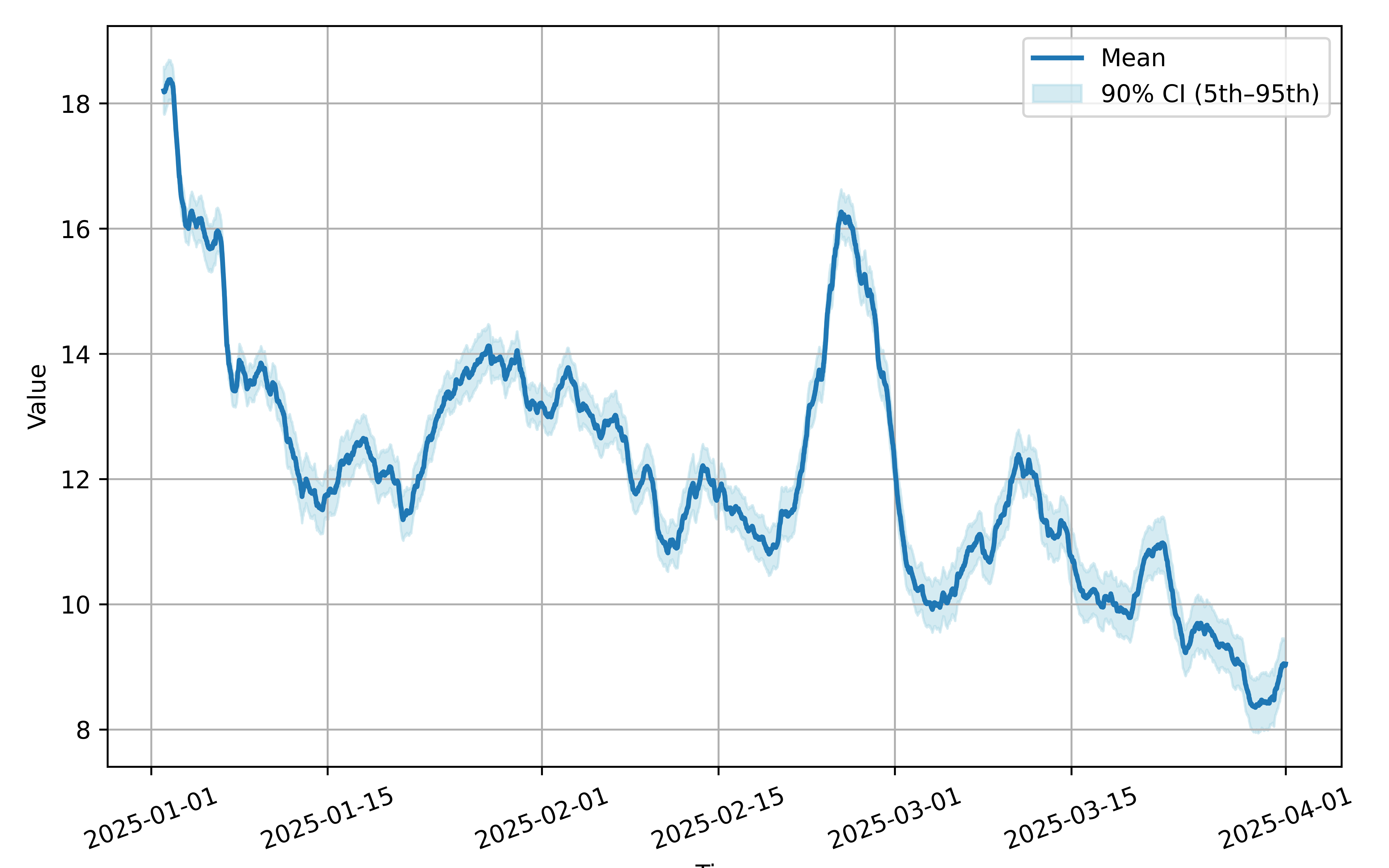}
    \caption{Moving Average of 90\% Prediction Interval on test data using CBB algorithm with GBR point estimate model.}
    \label{fig:rollingdayinterval}
\end{figure}

Figure \ref{fig:rollingdayinterval} shows the effectiveness of the CBB algorithm. A one-day moving average of the prediction interval at $90\%$ confidence interval with GBR point estimate model is plotted along with the observed values of the electricity demand on the test set.

\subsection{Comparative analysis}

\begin{figure}[h!]
    \centering
    \includegraphics[scale=0.85]{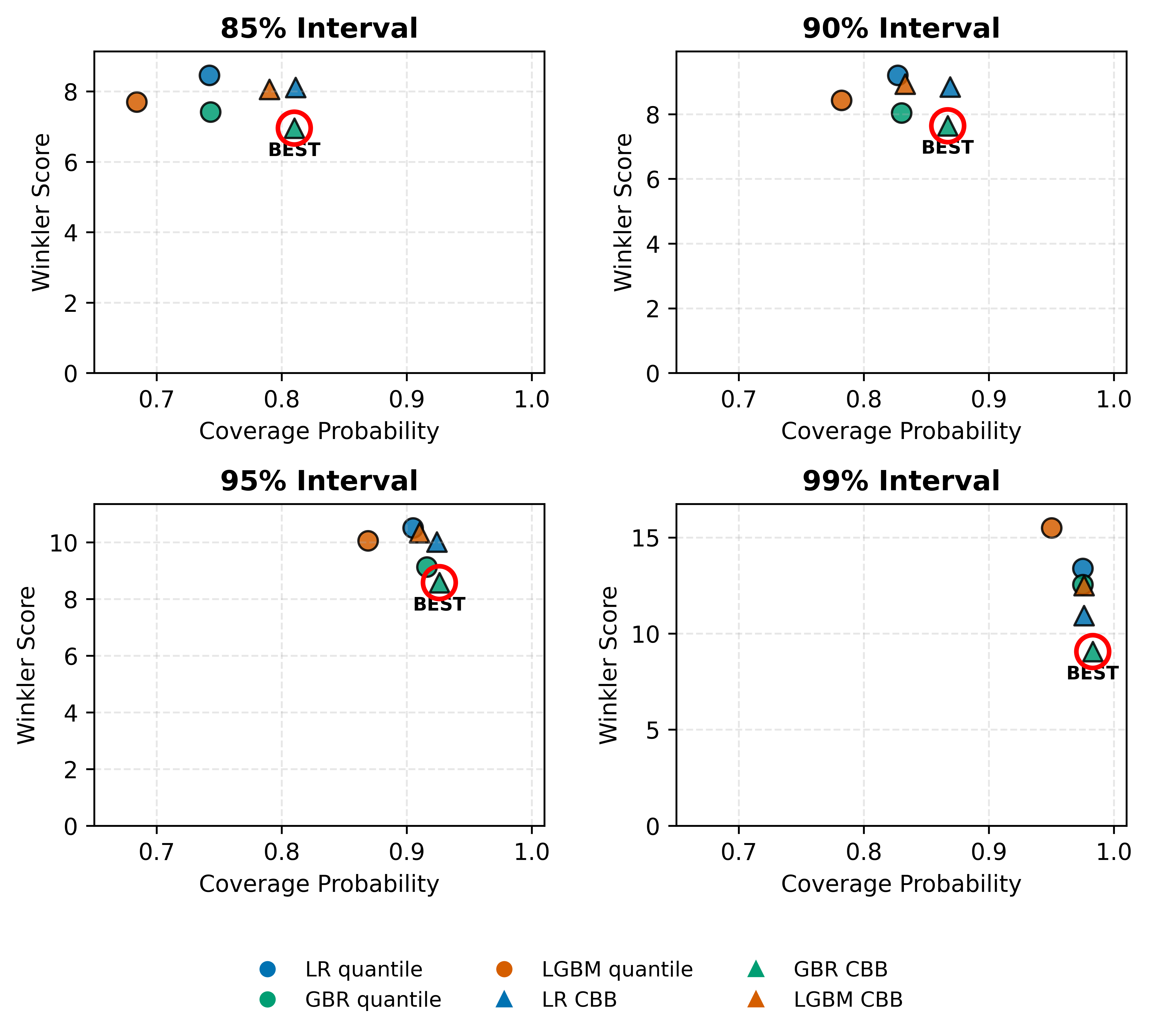}
    \caption{$CP$ vs $WS$ plots of (a) 85\% , (b) 90\% ,(c) 95\% and (d) 99\% confidence intervals for combinations of ML models and interval estimation algorithm}
    \label{fig:comparison scatter}
\end{figure}

The comparison of ensemble-based quantile regression and CBB is shown in the Figure \ref{fig:comparison scatter}. We drop block bootstrapping and Prophet as they have very high \( WS \). For each confidence interval size $(1-\alpha)100\%$, the values of $CP$ are plotted against $WS$. We summarize the comparison as follows:

\begin{enumerate}
  \item \textbf{Block Bootstrapping trades speed for quality.}  
        Averaged over the three point-estimate models, Block Bootstrapping trains
        ${\sim}94\%$ faster than ensemble-based Bootstrap Aggregating 
        (14.8 s vs.\ 229.7 s), but its prediction–interval width
        (as measured by \(WS\) at the 90 \% level) is
        5.6 \% larger, indicating a modest loss in sharpness.

  \item \textbf{CBB narrows intervals without a large runtime penalty.}  
        Relative to Block Bootstrapping, CBB lowers \(WS\) by 6.3 \% (90 \% level) while increasing training time by only about 32 \%. Compared to Bootstrap Aggregating, CBB’s intervals are 1.0 \% sharper yet the method remains more than 90 \% faster to train.

  \item \textbf{Clustering also improves coverage.}  
        CBB raises \(CP\) by 8.4 \% over Block Bootstrapping
        and by 5.3 \% over Bootstrap Aggregating at the 90 \% level,
        confirming that the homogeneous-residual sampling strategy boosts both sharpness \emph{and} reliability.

  \item \textbf{CBB with a GBR point model is the overall front runner.}  
        Across all confidence levels, the CBB\,+\,GBR combination delivers the narrowest intervals (average \(WS\) 22.6 \% lower than the ensemble-based LGBM benchmark) and the highest average coverage (9 \% higher than LGBM). Although LGBM Bootstrap Aggregating remains competitive on \(WS\) at the lower levels, it trails CBB\,+\,GBR on \(CP\) and requires far more compute.

  \item \textbf{Prophet produces the widest—hence least informative intervals.}  
        Its \(WS\) scores are an order of magnitude higher than the other methods, and while Prophet matches CBB’s coverage only at the very widest (95 - 99 \%) levels, it under-covers at the narrower levels and never attains comparable sharpness.
\end{enumerate}

The selection of the best model showed in Figure \ref{fig:comparison scatter} is based on a combined scoring approach that evaluates both \(CP\) and \(WS\). \(CP\) measures the reliability of prediction intervals, with higher values indicating better coverage. \(WS\), on the other hand, penalizes intervals that are either too wide or fail to capture the true values, with lower scores representing better quality. To balance these metrics, a normalized scoring formula is used:
\[
\text{Score} = 0.5 \times \text{Normalized \(CP\)} + 0.5 \times (1 - \text{Normalized \(WS\)}),
\]
where Normalized \(CP\) and Normalized \(WS\) are calculated as:
\[
\text{Normalized \textit{CP}} = \frac{CP - \min(CP)}{\max(CP) - \min(CP)},
\]
\[
\text{Normalized \textit{WS}} = \frac{WS - \min(WS)}{\max(WS) - \min(WS)}.
\]
This normalization ensures that both metrics are scaled between 0 and 1, enabling a fair comparison across models. The weights of $0.5$ for CP and $0.4$ for \( WS \) reflects the equal importance of reliability interval width in this study. The model with the highest combined score is identified as the best-performing model for each confidence level.

The analysis suggests that the CBB algorithm achieves comparable coverage as compared to the ensemble-based models while significantly improving the quality of prediction intervals. The CBB algorithm is based on the residual bootstrapping approach, thus also enjoys lower computation time as only one point estimate model is needed.

\section{Discussions and Future Work}
\label{sec:future_work}
In this work, we utilized an NN-based spectral clustering algorithm to group similar days of electricity demand and their residuals. By bootstrapping residual estimates in blocks from the closest cluster to the prediction day, we achieved improved prediction quality and comparable coverage probabilities against ensemble-based methods, while also reducing computation time. This efficiency is attributed to bootstrapping residual errors and adding them to point estimates, rather than relying on ensemble-based bootstrapping. 

The CBB approach demonstrated superior performance compared to Block Bootstrapping due to the incorporation of clustering. However, it is sensitive to the point estimate model, as residuals are obtained from the point forecasts, necessitating frequent retraining to ensure that the point forecasts model is correct and residuals are accurate. Although we selected the best ML models from the literature for point estimation, our primary focus was on constructing prediction intervals. Future research could investigate the performance of CBB with a wider variety of point estimate models, including non-ML-based ones like temporal fusion transformers (\cite{lim2021temporal}), which have shown higher accuracy in time-series forecasting.

Furthermore, our study only considered four residual clusters based on electricity demand. Exploring more complex clustering patterns, such as those based on the time of day or type of residence, could further enhance prediction intervals. Incorporating additional features, such as temperature values, and employing advanced clustering algorithms like density-based or fuzzy clustering (\cite{d2023wavelet}), may better capture the complex patterns in the data.

In summary, while the CBB approach outperforms ensemble-based methods, future research should aim to enhance point estimate models, incorporate exogenous variables, and refine the clustering process to achieve even more accurate and efficient forecasting.

\section{Conclusion}
\label{sec:conclusion}
This research focused on residual bootstrapping for constructing prediction intervals, highlighting the importance of uncorrelated residuals and constant variance. To meet these requirements, we proposed the CBB method that samples contiguous blocks of residuals and uses a spectral clustering scheme to ensure constant variance.

The residual bootstrapping approach aimed to reduce computation time while enhancing prediction accuracy, which it successfully achieved. The CBB algorithm demonstrated faster training times and produced narrower but better quality prediction intervals compared to ensemble methods, which, although having slightly better coverage, were slower due to training multiple weak learners.

The enhancement in prediction quality, attributed to the clustering scheme is evident from the \(WS\) scores of CBB compared to the Block Bootstrapping method, where clustering is the differentiating factor between the two approaches. As observed in Section \ref{sec: results}, clustering results in narrower interval widths compared to residual bootstrapping without clustering, thereby improving the quality of intervals.

This research suggested that clustering residuals before sampling them could enhance prediction interval accuracy, providing a foundation for developing more precise and efficient forecasting models. Future work may explore the effects of different point estimate models, not limited to ML methods, and improved clustering approaches on the CBB method.

\if0\blind{
\section*{Acknowledgements}
\noindent
The authors gratefully acknowledge funding from Triad National Security LLC under the grant from the Department of Energy National Nuclear Security Administration (award no. 89233218CNA000001).
} \fi

\bibliographystyle{chicago}
\spacingset{1}
\bibliography{elpibib2}
	
\end{document}